\documentclass{article}

\usepackage[nonatbib, preprint]{neurips_2024}
\usepackage[numbers]{natbib}

\usepackage[utf8]{inputenc} %
\usepackage[T1]{fontenc}    %
\usepackage[pagebackref,breaklinks,colorlinks,citecolor=Green]{hyperref}
\usepackage{url}            %
\usepackage{booktabs}       %
\usepackage{amsfonts}       %
\usepackage{nicefrac}       %
\usepackage{microtype}      %
\usepackage{microtype}
\usepackage{graphicx}
\usepackage[dvipsnames]{xcolor}
\usepackage{booktabs}
\usepackage{multirow}
\usepackage{listings}
\usepackage{amsmath}
\usepackage{amssymb}
\usepackage{mathtools}
\usepackage{amsthm}

\usepackage[capitalize,noabbrev]{cleveref}

\definecolor{faster_color}{HTML}{4FD1FF}
\definecolor{higher_color}{HTML}{00B050}
\definecolor{stronger_color}{HTML}{A875DD}

\theoremstyle{plain}

\theoremstyle{definition}

\theoremstyle{remark}

\newlength\savewidth\newcommand\shline{\noalign{\global\savewidth\arrayrulewidth
		\global\arrayrulewidth 1pt}\hline\noalign{\global\arrayrulewidth\savewidth}}
\newcommand{\tablestyle}[2]{\setlength{\tabcolsep}{#1}\renewcommand{\arraystretch}{#2}\centering\footnotesize}
\renewcommand{\paragraph}[1]{\vspace{1.25mm}\noindent\textbf{#1}}

\usepackage{xspace}
\makeatletter
\DeclareRobustCommand\onedot{\futurelet\@let@token\@onedot}
\def\@onedot{\ifx\@let@token.\else.\null\fi\xspace}
\def\eg{\emph{e.g}\onedot} 
\def\ie{\emph{i.e}\onedot}

\g@addto@macro{\endtabular}{\rowfont{}}%
\makeatother
\newcommand{\rowfonttype}{}%
\newcommand{\rowfont}[1]{%
\gdef\rowfonttype{#1}#1\ignorespaces%
}

\definecolor{hr}{gray}{0.7}  %
\definecolor{dt}{HTML}{ADCAD8}  %

\usepackage{array}
\newcolumntype{*}{>{\global\let\currentrowstyle\relax}}
\newcolumntype{^}{>{\currentrowstyle}}

\newcolumntype{H}{>{\setbox0=\hbox\bgroup}c<{\egroup}@{}}
\newcolumntype{Z}{>{\setbox0=\hbox\bgroup}c<{\egroup}@{\hspace*{-\tabcolsep}}}

\usepackage{subfig}
\usepackage{graphicx}
\makeatother

\usepackage{pifont}
\newcommand{\cmark}{\ding{51}}
\newcommand{\xmark}{\ding{55}}

\usepackage[textsize=tiny]{todonotes}
\begin{document}

\title{PaLM2-VAdapter: \\Progressively Aligned Language Model Makes\\ a Strong Vision-language Adapter}

\author{
    Junfei Xiao\textsuperscript{1,2},
    Zheng Xu\textsuperscript{2},
    Alan Yuille\textsuperscript{1},
    Shen Yan\textsuperscript{2\dag},
    Boyu Wang\textsuperscript{2\dag} \\
    \\
    \textsuperscript{1}Johns Hopkins University \qquad
    \textsuperscript{2}Google Research
}

\maketitle

\begin{abstract}
This paper demonstrates that a progressively aligned language model can effectively bridge frozen vision encoders and large language models (LLMs). While the fundamental architecture and pre-training methods of vision encoders and LLMs have been extensively studied, the architecture and training strategy of vision-language adapters vary significantly across recent works. Our research undertakes a thorough exploration of the state-of-the-art perceiver resampler architecture and builds a strong baseline. However, we observe that the vision-language alignment with perceiver resampler exhibits slow convergence and limited scalability with a lack of direct supervision. To address this issue, we propose \textbf{PaLM2-VAdapter}, employing a progressively aligned language model as the vision-language adapter. Compared to the strong baseline with perceiver resampler, our method empirically shows faster convergence, higher performance and stronger scalability.
Extensive experiments on various Visual Question Answering (VQA) and captioning tasks for both images and videos demonstrate that our model exhibits state-of-the-art visual understanding and multi-modal reasoning capabilities. Notably, our method achieves these advancements with  30$\sim$70\% fewer parameters than the state-of-the-art large vision-language models, marking a significant efficiency improvement.
\end{abstract}

\vspace{-3mm}
\section{Introduction}

With the notable successes of  large language model (LLM) ~\cite{gpt-3,touvron2023llama,palm2}, coupled with advancements in vision-language pretraining~\cite{clip,align,blip,coca}, researchers are now well-equipped to construct sophisticated Large Vision-Language Models (LVLMs).
This is achieved by integrating robust unimodal models, namely vision encoders and LLMs, thereby circumventing the need to develop these models from scratch~\cite{flamingo, blip-2, llava,pali}. These LVLMs have demonstrated exceptional performance across a variety of multi-modal benchmarks, showcasing their impressive capabilities in understanding, reasoning, and generalizing across different contexts~\cite{flamingo,blip-2,anymal}.

\begin{figure}[h]
    \centering
    \includegraphics[width=0.7\linewidth]{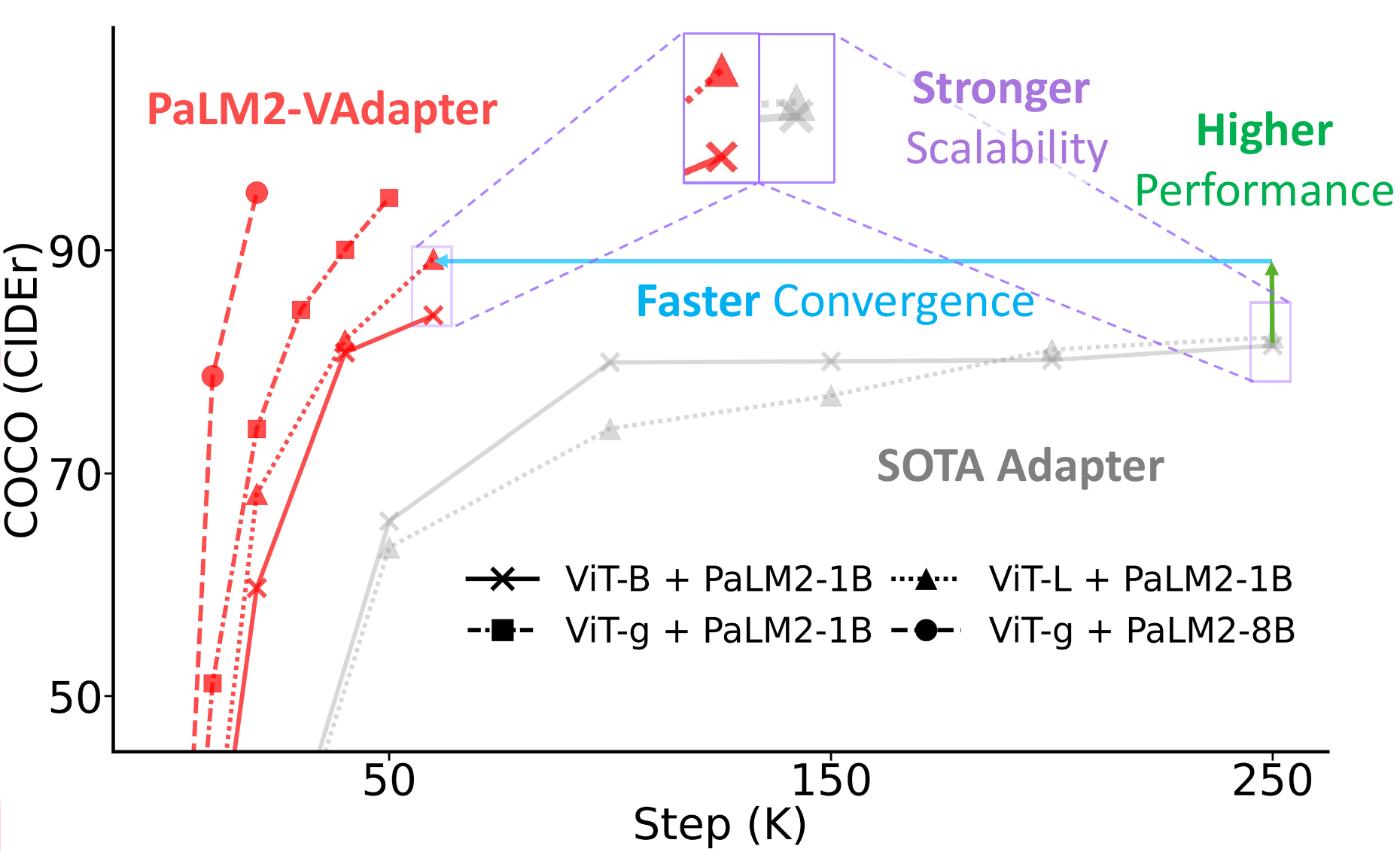}
     
    \caption{\textbf{\textcolor{faster_color}{Faster}, \textcolor{higher_color}{higher}, and \textcolor{stronger_color}{stronger}.} Our progressively aligned language model demonstrates faster convergence, higher performance and stronger scalability as an adapter for vision-language alignment.}
    \label{fig:teaser_figure}
    \vspace{-3mm}
\end{figure}

 Contrasting with traditional full-model finetuning approaches, recent research has shifted towards freezing both vision encoder and LLM during LVLM training~\cite{flamingo,blip-2,anymal}. There are two main reasons for this. Firstly, vision encoders and LLMs have learned very strong feature extraction ability and reasoning ability through the large-scale pretraining on high-quality data, and finetuning could lead to catastrophic forgetting. Secondly, as these base models are getting bigger, freezing them saves training costs. Therefore, the focus is on training an adapter that connects the vision encoder and the LLM for cross-modality alignment.

To build strong LVLMs using pre-trained and frozen vision encoders and LLMs, the keys lie in the design and training strategy of the adapter. Existing methods like Flamingo and AnyMAL~\cite{flamingo, anymal} employ the perceiver resampler as their adapter architecture, resulting an effective way for cross-modality alignment. On the other hand, BLIP-2~\cite{blip-2} tackles the adapter pre-training issue by introducing Q-Former, which takes an additional pretraining stage with multi-task learning on image-text pairs. Although these methods demonstrate impressive performance, questions regarding the optimal adapter architecture and the necessity of adapter pretraining still remain open for exploration.

To address the open questions in the design and training of adapters for LVLMs, we conduct an in-depth study into the latest cross-attention based adapter architectures, particularly focusing on the perceiver resampler and make a strong baseline. However, we observed that the perceiver resampler adapter exhibits slow convergence and limited scalability, especially when scaling up the vision encoder. To overcome these challenges, we propose \textbf{PaLM2-VAdapter}, which employs a progressive alignment strategy for bridging frozen vision encoders and LLM decoders. 
Specifically, the classic alignment framework is used in a progressive way with two stages and a tiny PaLM-2 model is trained as different roles (stage 1: LM decoder, stage 2: adapter). Compared to the baseline models using state-of-the-art adapters, \textbf{PaLM2-VAdapter} demonstrates  faster convergence, higher performance and stronger scalability, as detailed in \cref{fig:teaser_figure}.

We evaluate our models on various vision-language benchmarks in both image-based and video-based captioning and QA tasks. Our models consistently show state-of-the-art or comparable performance, while only requiring 30$\sim$80\% fewer parameters than previous models. This efficiency underscores the effectiveness of our proposed \textbf{PaLM2-VAdapter} in advancing the field of LVLMs.

 To sum up, our contributions lie in three folds:
    \begin{enumerate}
        \item We conduct a comprehensive study of the state-of-the-art adapter architecture (\ie, perceiver resampler) and build a strong baseline with it.
        \item We propose \textbf{PaLM2-VAdapter}, a progressive alignment strategy to train a tiny PaLM2 language model as the vision-language adapter, making solid improvement on convergence, performance and scalability.

        \item Our models achieve state-of-the-art performance on various visual captioning and QA benchmarks but use 30$\sim$70\% less parameters than other models.
    \end{enumerate}

\vspace{-3mm}
\section{Related Work}

\vspace{-3mm}
\subsection{Vision-language Pre-training}
Vision-language pre-training aims to learn universal multimodal representations through a set of pretraining objectives, including image-text matching~\cite{ablef, bao2022vlmo, dou2022coarse}, image-text contrastive learning~\cite{clip, align, yang2022vision, duan2022multi}, and also auto-regressive image captioning~\cite{blip, coca, wang2021ufo, wang2021simvlm}. However, models pretrained on image-text pairs often lack the complex reasoning and few-shot learning abilities of Large Language Models (LLMs), primarily due to their focus on image captions~\cite{coco, clip, align, laion, wit}. To overcome this, recent efforts have shifted towards integrating pretrained vision encoders and LLMs into larger vision-language models. This strategy extends their capabilities to more advanced tasks such as image captioning and Visual Question Answering (VQA), leveraging LLMs for improved performance.

\subsection{Large Language Models (LLMs)}
Arming with scaled-up data and models, Large Language Models (LLMs) have demonstrated emergent capabilities like zero-shot generalization and in-context learning ability. This has sparked a surge in research and development, leading to significant advancements in models like FlanT5~\cite{flant5}, PaLM 2~\cite{palm2}, GPT-4~\cite{gpt4}, LLaMA~\cite{touvron2023llama} and etc. Given the complex reasoning and remarkable understanding ability, LLMs are utilized as a "head". In this paper, we aims to bridge strong vision encoders with the PaLM 2 series of LLMs, extending its capability to understand and do reasoning with visual embeddings. To avoid the PaLM 2 model losing any knowledge or its strong language reasoning ability, our method keeps the large PaLM 2 model frozen all the time.
\vspace{-3mm}
\subsection{Large Vision-language Models (LVLMs)}

Large Vision-language Models (LVLMs) connect vision and language together and extend the reasoning ability of LLMs to process with multi modal input. Numerous works have been proposed in this direction, including Flamingo~\cite{flamingo}, OpenFlamingo~\cite{openflamingo}, BLIP-2~\cite{blip-2}, InstructBLIP~\cite{instructblip}, MiniGPT-4~\cite{minigpt4}, LLaVA~\cite{llava}. Flamingo is the first work in this line, which uses the perceiver resampler as an adapter to feed visual tokens into language models. However, the number of trainable parameters in Flamingo is still more than billions, making the alignment with limited efficiency. BLIP-2 proposes a lightweight Q-Former as the adapter. However, the Q-Former needs a complex training process, including a two-stage training with three training objectives (vision-lanauge contrastive loss, matching loss and generation loss). InstructBLIP and MiniGPT-4 are extensions of BLIP-2 by using instruction tuning data or additional projection layer. LLaVA uses a simple projection layer to convert vision representations into the same dimension as the language. In this paper, we propose a progressive alignment strategy to use a pre-trained language model as the adapter, which shows faster convergence, higher performance and stronger scalability than the state-of-the-art perceiver resampler.

\vspace{-3mm}
\section{Method}
\label{sec:smethod}
\vspace{-3mm}

Our study is based on a classic visual-language alignment pipeline which keeps the visual encoder and large language model (LLM) frozen all the time. An adapter is inserted between the vision encoder and LLM to project the encoded visual embeddings to the language representation space. This section firstly provides a preliminary overview of vision-language adapter architectures  (\S\ref{sec:method_adapter_arch}) and then explains the model  framework of visual-language alignment with adapter (\S\ref{sec:method_alignment_with_adapter}). Lastly, we present our method using progressive vision-language alignment strategy for training a tiny language model as adapter (\S\ref{sec:method_llm_as_adapter}).

\begin{figure*}[t]
    \centering
    \includegraphics[width=0.96\linewidth]{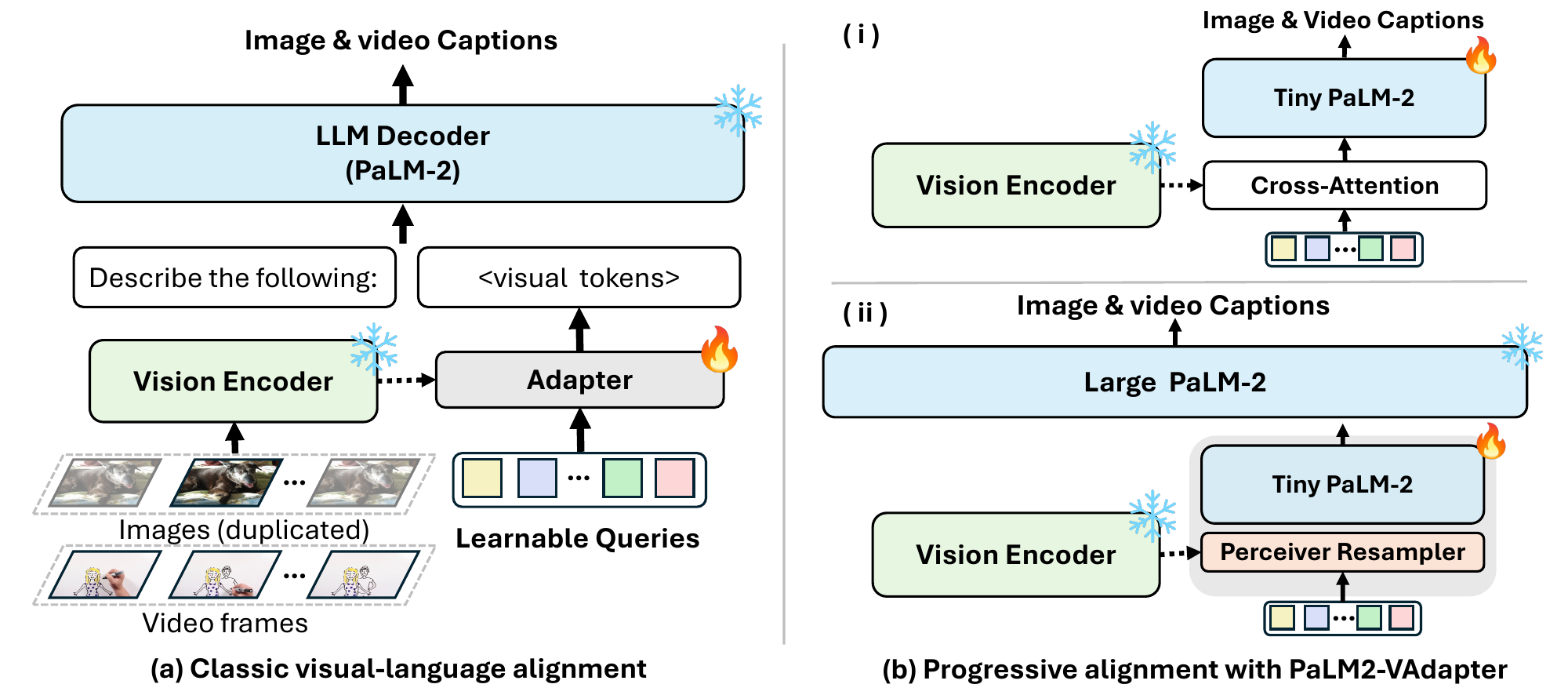} 
     
    \caption{\textbf{Method overview.}  \textbf{(a): }The classic model framework for visual-language alignment, consisting of three major parts: a vision encoder, an adapter and a LLM decoder. \textbf{(b): }Our progressive alignment strategy of our PaLM2-VAdapter. (i) A tiny PaLM2 language model ($\sim$108M) is trained as the LM decoder in the first stage and (ii) then  trained as the vision-language adapter (with an addition 1-layer perceiver resampler) for aligning the same vision encoder and a large PaLM2 decoder.}
    \vspace{-3mm}
    \label{fig:method_overview}
    
\end{figure*}
\vspace{-3mm}
\subsection{Preliminary}
\label{sec:method_adapter_arch}
Existing large vision-language models adopt various kinds of adapter architectures for cross-modality alignment. In this paper, we present an in-depth exploration of the state-of-the-art cross-attention based adapters and propose to progressively aligned self-attention based language model.

\paragraph{Cross-attention based adapter.}
The adapters in this style adopt the cross-attention mechanism for visual feature alignment. Specifically, the visual features extracted by the vision encoder are served as the keys and values which are cross-attentioned to a set of learnable queries, shown in \cref{fig:method_overview}a. We conduct a comprehensive study of the state-of-the-art perceiver resampler architecture and establish a very strong baseline model using 6-layer perceiver resampler as the adapter (detailed in \S\ref{sec:strong_baseline_resampler}).

\paragraph{Self-attention based adapter.} Self-attention layers can also be introduced in adapters to improve representation quality. Notably, self-attention based adapters could use pretrained language models for initialization to get better convergence and improve the performance.

\vspace{-3mm}
\subsection{Visual-language Alignment with Adapter}
\label{sec:method_alignment_with_adapter}
As shown in \cref{fig:method_overview}a, the vision-language model has three major parts: vision encoder, visual adapter and LLM.  The target is to align the visual features with the LLM representation space. The visual encoder and the LLM are both frozen all the time. This setup greatly reduces training cost and preserves their strong visual feature extraction and reasoning ability learned from large-scale pre-training.  The CoCa~\cite{coca} vision encoder is pre-trained with image-text pairs and is used to convert images and video frames into a set of feature tokens. These feature tokens are projected by a lightweight visual adapter to the LLM representation space. We adopt PaLM 2~\cite{palm2} series models as the LLM decoder and the training task is to generate captions based on the visual embedded prefix.

\vspace{-3mm}
\subsection{Progressive Visual-language Alignment}
\label{sec:method_llm_as_adapter}

As language models emerge strong  representation ability through the generative pre-training task and usually shows great scalability, we propose to introduce a tiny PaLM2 language model, using a progressive vision-language alignment strategy to make strong vision-language adapters. Specifically, our method uses a tiny PaLM2 language model (TLM) as the adapter and trains it in a progressive way, which consists of two stages:

\noindent\textbf{Stage 1 - TLM trained as the decoder:} In the first stage, the language model starts from a pretrained tiny PaLM2 model ($\sim$108M) and is finetuned with the classic vision-language alignment task ( shown in \cref{fig:method_overview}b(i)). 

\noindent\textbf{Stage 2 - TLM trained as the adapter:} In the second stage, given this pre-aligned tiny PaLM2 model, an additional 1-layer perceiver resampler is added before the aligned tiny PaLM2 model to bridge the same vision encoder and a larger PaLM2 model (shown in \cref{fig:method_overview}b(ii)).  

Compared to our strongest model with state-of-the-art adapter (\ie, perceiver resampler), our method is proven to have faster convergence, higher performance and stronger scalability (detailed in \S\ref{sec:faster_higher_stronger}). In addition to the effective architecture, the proposed progressive alignment strategy greatly advance PaLM2-VAdapter, making remarkable improvements for vision-language alignment (detailed in \S\ref{sec:progressively_pretrain_adapter}). Notably, the additional perceiver resampler is very crucial for efficient cross-modality fusion based on our empirical observation (detailed in \S\ref{sec:perceiver_is_needed}).

\vspace{-3mm}
\section{Experiments}
\label{sec:experiments}
\vspace{-3mm}

\subsection{Implementation Details}

\noindent\textbf{Model.} We adopt CoCa~\cite{coca} pretrained ViTs as our vision encoders. The input resolution is 288 and the patch size is 18x18. We adopt PaLM 2~\cite{palm2} pretrained models as the LLM. Perceiver resampler~\cite{flamingo} is used as the baseline adapter architecture, with 256 learnable queries. Our proposed adapter consists of a 1-layer perceiver resampler and a tiny transformer-based language model ($\sim$110M). Two fully-connected layers are applied before and after adapter for dimension matching.

\noindent\textbf{Data.}
Our models are trained on image-text paired data of WebLI~\cite{pali} dataset and video-text paired data of VTP~\cite{flamingo} and SMIT~\cite{smit} datasets. The ablations in \cref{tab:ablation_cross_attention_adapter} are solely trained on WebLI.

\noindent\textbf{Training.}
The images and videos are duplicated or sampled to 8 frames~\cite{videococa} as the visual inputs. The base learning rate is 5e-4 and is scheduled with a  warm-up and linear decay. The training batch size is 2048. By default, experiments are trained with 250K steps. We use a prompt template of "Describe the following: $<$visual tokens$>$" for training. For detailed information, please refer to \cref{sec:imple_details}.

\noindent\textbf{Evaluation.}
All the input resolution is the same as training (\ie, 288) with a patch size of 18. We evaluate our method on captioning tasks and Visual Question Answering (VQA) tasks for both images and videos.  Specifically, COCO~\cite{cococap}, VQAv2~\cite{vqav2}, TextVQA~\cite{textvqa}, VizWiz~\cite{bigham2010vizwiz}, OKVQA~\cite{okvqa} are used for image-based evaluation. MSRVTT~\cite{msrvtt}, VATEX~\cite{wang2019vatex}, MSVD-QA~\cite{msvd_qa}, and iVQA~\cite{yang2021justask} are used for video-based evaluation. We use different prompts for the LLM decoder on different tasks. For detailed prompts information, please refer to \cref{sec:imple_details}\&\ref{sec:zero_shot_vqa}.

\begin{table*}[t]
\centering

\subfloat[\textbf{LayerNorm options.}]{
    \tablestyle{1pt}{1.1}
    \begin{tabular}{cc|cc}
    {Query   \& Key } &  {Final} & COCO Cap. & VQAv2 (Val) \\\shline
    \xmark              & \cmark     & 38.4               & 32.2                \\
    Shared                           & \xmark     & 44.0               & 46.7                \\
    Separate                             & \xmark     & \textbf{46.8}      & \textbf{52.5}       \\
    Separate                             & \cmark     & 36.2               & 37.6               
    \end{tabular}
}
\subfloat[\textbf{Feed-forward network \& time embedding.}]{
    \tablestyle{2pt}{1.5}
    \begin{tabular}{cc|cc}
    {FFN} & {Time embedding} & COCO Cap. & VQAv2 (Val) \\\shline
    \cmark               & \xmark                          & 34                 & 38.3                \\
    \xmark               & \cmark                          & 33.8               & 45.1                \\
    \cmark               & \cmark                          & \textbf{46.8}      & \textbf{52.5}      
    \end{tabular}
}

\subfloat[\textbf{Query dimension.}]{
\tablestyle{1pt}{1.2}
\begin{tabular}{c|cc}
 {Query   Dim} & COCO Cap. & VQAv2\\   \shline
384                          & 40.9               & 45.4                \\
768                          & \textbf{46.8}      & \textbf{52.5}       \\
1536                         & 38.3               & 45.0               
\end{tabular}
}
\subfloat[\textbf{Hidden dimension.}]{
\tablestyle{1pt}{1.2}
\begin{tabular}{c|cc}
Hidden   Dim & COCO Cap. & VQAv2 \\ \shline
384          & 40.6               & 46.7               \\
768          & \textbf{46.8}      & \textbf{52.5}      \\
1536         & 38.5               & 32.1              
\end{tabular}
}
\subfloat[\textbf{Number of layers.}]{
\tablestyle{1pt}{1.2}
\begin{tabular}{c|cc}
\#Layers & COCO Cap. & VQAv2 \\ \shline
1        & 37.7               & 37.5               \\
3        & 40.8               & 47.6               \\
6        & \textbf{46.8}      & \textbf{52.5}     
\end{tabular}
}
 
\caption{\textbf{In-depth analysis with key components of perceiver resampler.} Results on COCO captioning benchmark (CIDEr score) and VQAv2 validation set (accuracy) are reported.  Models are trained on WebLI (image-text paired dataset).}
\vspace{-3mm}
\label{tab:ablation_cross_attention_adapter}
\end{table*}
\vspace{-2mm}
\subsection{A Strong Baseline with Perceiver Resampler}
\vspace{-2mm}
\label{sec:strong_baseline_resampler}
To figure out the effectiveness of different model components of cross-attention based adapters , we conduct a comprehensive ablation study based on perceiver resampler, which is the state-of-the-art adapter architecture. As shown in \cref{tab:ablation_cross_attention_adapter}, our study covers different choices to apply LayerNorm, important modules (\ie, Feed-Forward Network FFN and time embedding), dimension of queries and cross-attention layers and also the number of perceiver resampler layers. 

Based on the empirical results, we get several design rules for perceiver resampler based adapter: 1) LayerNorms are important and should be separately applied to the queries and the cross-modality inputs (as keys and values). 2) Feed-Forward Network (FFN) and time embedding make the adapter training stable and effective and can greatly improve the performance. 3) The dimension of the learnable queries and the cross-attention layer should be set moderate. Following this rules, we build a very strong baseline achieving 81.4 CIDEr on COCO captioning, 38.2 CIDEr on MSRVTT captioning and 53.1 accuracy on VQAv2.

\vspace{-3mm}

\subsection{Faster, Higher, and Stronger}
\label{sec:faster_higher_stronger}

\begin{minipage}[t]{0.4\textwidth}
\vspace{-5ex}
Although the baseline shows reasonable performance, we observe that it has limited scalability and slow convergence (shown in \cref{fig:teaser_figure}). To address these issues, we propose to introduce a tiny language model as an adapter and train it progressively (shown in \cref{fig:method_overview}b). Compared to the strong baseline based on state-of-the-art architecture (shown in \cref{tab:faster_higher_stronger_tab}), our PaLM2-VAdapter shows:
\end{minipage}%
\hfill%
\begin{minipage}[t]{0.58\textwidth}
\centering
\tablestyle{1pt}{1.1}
\footnotesize
\begin{tabular}{cc|ccc}
\multirow{2}{*}{Method} & \multirow{2}{*}{Vision Enc.} & Converg. & \textbf{COCO} & \textbf{MSRVTT} \\ 
& & Steps (K) & CIDEr & CIDEr \\ \hline
Perceiver Res. & ViT-B & 250 & 81.4 & 38.2 \\
PaLM2-VAdapter & ViT-B & \textbf{60} \textcolor{red}{(-76\%)} & \textbf{83.0} \textcolor{Green}{(+1.6)} & \textbf{42.1} \textcolor{Green}{(+3.9)} \\ \hline
Perceiver Res. & ViT-L & 250 & 82.4 & 38.2 \\
PaLM2-VAdapter & ViT-L & \textbf{60} \textcolor{red}{(-76\%)} & \textbf{89.6} \textcolor{Green}{(+7.2)} & \textbf{42.7} \textcolor{Green}{(+4.5)}
\end{tabular}
\captionof{table}{\textbf{Faster, higher and stronger.} Compared to the perceiver resampler baseline, \textbf{PaLM2-VAdapter} shows faster convergence, higher performance and stronger scalability. PaLM2-1B is used as the LLM decoder.}
\label{tab:faster_higher_stronger_tab}
\end{minipage}

\noindent\textbf{Faster convergence.} While the perceiver resampler baselines take 250K steps to converge, our PaLM2-VAdapter only need 60K steps to converge which is $\sim$3$\times$faster.

\noindent\textbf{Higher performance.} PaLM2-VAdapter achieves much higher performance than the baseline perceiver resampler models (ViT-B: 83.0 vs. 81.4,  ViT-L: 89.6 vs. 82.4) when aligning the same vision encoder and LLM decoder pairs.

\noindent\textbf{Stronger scalability.} Perceiver resampler shows marginal improvement when the vision encoder is scaled from ViT-B to ViT-L. However, our PaLM2-VAdapter makes much larger improvement (COCO: 6.6 vs 1.0, MSRVTT: 0.6 vs 0.0) , showing stronger scalability.

\vspace{-3mm}
\subsection{Progressive Training Does Help}
\label{sec:progressively_pretrain_adapter}

We conduct a comparison regarding different pre-training strategies using the same adapter architecture (1-layer perceiver resampler + PaLM2-108M), detailed in \cref{tab:adapter_pretrain_tasks}. The ablation compares three training strategies for the adapter: a) randomly initialized; b) Generative pre-trained on language data (PaLM2 pretraining) , initialized from a PaLM2 checkpoint; c) Pretrained with the proposed progressive training strategy. The tiny PaLM2 model is first initialized from the PaLM2 checkpoint and then fine-tuned with vision-language generative pre-training (stage 1, the tiny PaLM2 model is trained as the LM decoder).   The results prove the effectiveness of the progressive training strategy applied to the adapter including language-only generative pre-training \cite{palm2} and 
vision-language generative pre-training (stage 1, shown in \cref{fig:method_overview}b(i)).

\begin{table}[h]
\centering
\begin{minipage}[t]{0.48\textwidth}
\tablestyle{1pt}{1.2}
\centering
\begin{tabular}{cc|cc}
Language Only         & Vision-language       & \textbf{COCO}                 & \textbf{VQAv2}                \\
 (LM pretraining)          &  (stage-1)             & CIDEr               & Accuracy             \\ \shline
\xmark               & \xmark                 & 79.2                 & 50.8                 \\
\cmark               & \xmark                 & 81.3                 & 52.1                 \\
\cmark               & \cmark                 & \textbf{83.0}                 & \textbf{53.8}     
\end{tabular}
\caption{\textbf{Comparison of different adapter pre-training settings.} Both language-only generative pre-training (PaLM2) and vision-language generative pre-training (stage-1, language model as decoder) can improve the final aligned large vision-language model's performance.}
\label{tab:adapter_pretrain_tasks}
\end{minipage}%
\hfill
\begin{minipage}[t]{0.48\textwidth}
\tablestyle{1pt}{1.2}
\centering
\begin{tabular}{cc|cc}
{Cross-attention} & \multirow{2}{*}{\# Layers} & \textbf{COCO}          & \textbf{VQAv2}        \\ Module Type
                                        &                           & CIDEr      & Accuracy    \\ \shline
Attentional Pooler                & 1                         & 81.1          & 53.5          \\
Perceiver Resampler                     & 1                         & \textbf{85.6} & \textbf{55.1} \\
Perceiver Resampler                     & 6                         & 70.3          & 49.7         
\end{tabular}
\caption{\textbf{Comparison of using different types of cross-attention modules.} A lightweight perceiver resampler cross-attention module is the best cross-modality fusion choice for PaLM2-VAdapter.}
\label{tab:resampler_is_needed}
\end{minipage}
\vspace{-3mm}
\end{table}

\noindent
\begin{table}[!t]
\begin{minipage}[t]{0.48\textwidth}
\centering
\tablestyle{1pt}{1.2}
\begin{tabular}{lccc}
\multirow{2}{*}{Method} &{\scriptsize \# Total} & {\scriptsize \# Trainable} & {\scriptsize \textbf{COCO}} \\
& {\scriptsize Params} & {\scriptsize Params} & {\scriptsize CIDEr} \\ \shline
CM3Leon\cite{yu2023scaling} & 7B & 7B & 61.6 \\
Flamingo-3B\cite{flamingo}& 3.2B & 1.2B & 73.0 \\
Flamingo-9B\cite{flamingo} & 9.3B & 1.6B & 79.4 \\
Flamingo-80B\cite{flamingo} & 80B & 10.2B & 84.3 \\
IDEFICS-9B\cite{IDEFICS} & 9B & 1.5B & 46.0 \\
IDEFICS-80B\cite{IDEFICS} & 80B & 14B & 91.8 \\
AnyMAL-15B\cite{anymal} & 15B & 100M$^*$ & \textbf{99.5} \\
\hline
PaLM2-VAdapter 1B (ViT-B) & 1.8B & 120M & 83.0 \\
PaLM2-VAdapter 1B (ViT-L) & 2.0B & 120M & 89.6 \\
PaLM2-VAdapter 1B (ViT-g) & 2.8B & 130M & \underline{97.5} \\
PaLM2-VAdapter 8B (ViT-g) & 10.8B & 130M & 95.2 \\
\end{tabular}
\caption{\textbf{Zero-shot Image Captioning.} The best result is \textbf{bolded} and the second-best result is \underline{underlined}. Our model demonstrates comparable zero-shot visual understanding ability. *: Estimated by given information.}
\label{tab:image_cap}
\end{minipage}%
\hfill%
\begin{minipage}[t]{0.48\textwidth}
\centering
\tablestyle{0.1pt}{1.45}
\begin{tabular}{lcc|cc}
\multirow{2}{*}{Method} & {\scriptsize \# Total} & {\scriptsize \# Trainable} & \multicolumn{1}{c}{\scriptsize \textbf{MSRVTT }} & \multicolumn{1}{c}{\scriptsize \textbf{VATEX }} \\
& {\scriptsize Params} & {\scriptsize Params} & {\scriptsize CIDEr} & {\scriptsize CIDEr} \\ \shline
VideoCoCa\cite{videococa} & 2.1B & 2.1B & 27.1 & 22.8 \\
DeCap\cite{li2023decap} & 140M & 50M & 34.8 & 18.7 \\
Flam.-3B\cite{flamingo} & 3.2B & 1.2B & - & 40.1 \\
Flam.-9B\cite{flamingo} & 9.3B & 1.6B & - & 39.5 \\
Flam.-80B\cite{flamingo} & 80B & 14B & - & 46.7 \\
\hline
PaLM2-VA.1B(ViT-B) & 1.8B & 120M & 42.1 & 38.3 \\
PaLM2-VA. 1B(ViT-L) & 2.0B & 120M & 42.7 & 45.5 \\
PaLM2-VA. 1B(ViT-g) & 2.8B & 130M & \underline{45.6} & \underline{51.2} \\
PaLM2-VA. 8B(ViT-g) & 10.8B & 130M & \textbf{47.7} & \textbf{53.0} \\ 
\end{tabular}
\caption{\textbf{Zero-shot Video Captioning.} The best result is \textbf{bolded} and the second-best result is \underline{underlined}. Our model demonstrates the state-of-the-art zero-shot visual understanding ability on videos.}
\label{tab:video_cap}
\end{minipage}
\vspace{-6mm}
\end{table}

\begin{figure*}[!h]
    \centering
    \includegraphics[width=\linewidth]{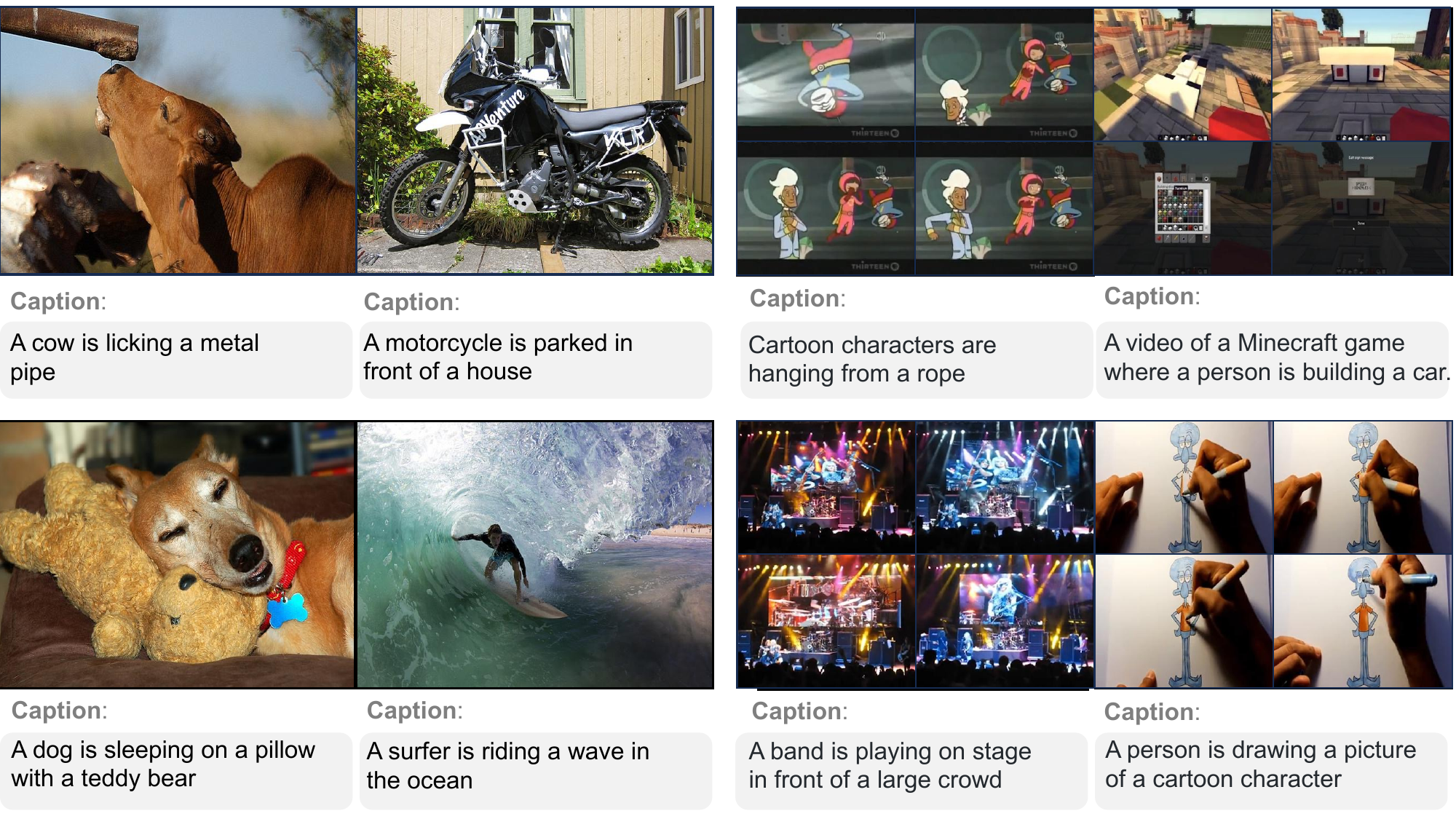} 
     
    \caption{\textbf{Qualitative examples of Visual Captioning.} \textbf{Left:} Image captioning on the COCO dataset. \textbf{Right:} Video captioning on the MSRVTT dataset. PaLM2-VAdapter demonstrates strong visual understanding ability. }
    \label{fig:examples_caption}
    
\end{figure*}

\begin{table*}[!h]
\centering
\tablestyle{5pt}{1.0}
\begin{tabular}{ccc|cccc}

\multirow{2}{*}{Method} & \# Total & \# Trainable & \textbf{VQAv2} & \textbf{TextVQA} & \textbf{VizWiz} & \textbf{OKVQA} \\
& Params & Params & Accuracy & Accuracy & Accuracy & Accuracy \\ \shline
Flamingo-3B \cite{flamingo} & 3.2B & 1.2B & 49.2 & 30.1 & 28.9 & 41.2 \\ 
Flamingo-9B \cite{flamingo} & 9.3B & 1.6B & 51.8 & 31.8 & 28.8 & 44.7 \\ 
Flamingo-80B \cite{flamingo} & 80B & 10.2B & 56.3 & \textbf{35.0} & 31.6 & \textbf{50.6} \\
BLIP-2 (FlanT5xxL) \cite{blip-2} & 12.1B & 108M & \textcolor{gray}{65.0}$^\dagger$ & \textcolor{gray}{44.1}$^*$ & 29.4 & \underline{45.9} \\
InstructBLIP (V-13B) \cite{instructblip}  & 14.1B & 108M & - & \textcolor{gray}{50.7}$^{\dagger*}$ & 33.4 & - \\

IDEFICS-9B \cite{IDEFICS} & 9B & 1.5B & 50.9 & 25.9 & 35.5 & 38.4 \\
IDEFICS-80B \cite{IDEFICS} & 80B & 14B & \underline{60.0} & \underline{30.9} & 36.0 & 45.2 \\
AnyMAL 13B (ViT-G) \cite{anymal} & 15B & 100M & 59.6 & 24.7 & 24.4 & 33.1\\\hline 
PaLM2-VAdapter 1B (ViT-B) & 1.8B & 120M & 53.8 & 18.7 & 28.6 &31.0\\
PaLM2-VAdapter 1B  (ViT-L) & 2.0B & 120M & 55.0  & 22.2 & 37.2 & 31.7\\
PaLM2-VAdapter 1B  (ViT-g) & 2.8B & 130M & 57.9  & 23.7 & \textbf{44.1} & 33.6\\
PaLM2-VAdapter 8B  (ViT-g) & 10.8B & 130M & \textbf{60.6} & 24.8  & \underline{43.7}  & 40.9\\

\end{tabular}
 
\caption{\textbf{Zero-shot Image Question Answering.}  The best result is \textbf{bolded} and the second-best result is \underline{underlined}. Our model demonstrates strong zero-shot vision-language reasoning ability on the four classic benchmarks, comparable to the state-of-the-art methods. *: with additional OCR inputs. $\dagger$ : in-domain images were used.}
\label{tab:image_qa}

\end{table*}

\begin{table*}[!ht]
    \centering
    \tablestyle{4pt}{1.1}
        \begin{tabular}{ccc|ccc}

\multirow{2}{*}{Method} & \# Total & \# Trainable  & \textbf{MSRVTT-QA} & \textbf{MSVD-QA} & \textbf{iVQA} \\
& Params & Params & (Top-1 Acc.) & (Top-1 Acc.) & (iVQA Acc.) \\
\shline
Just Ask \cite{yang2021justask} & 600M & 600M & 5.6 & 13.5 & 13.3 \\
HiTeA \cite{ye2023hitea} & 297M & 297M & 8.6 & 18.2 & - \\
FrozenBiLM \cite{yang2022frozenbilm} & 890M & 30M & 16.9 & 33.7 & 26.2 \\
Flamingo-3B \cite{flamingo} & 3.2B & 1.2B & 11.0 & 27.5 & 32.7 \\
Flamingo-9B \cite{flamingo} & 9.3B & 1.6B & 13.7 & 30.2 & {35.2} \\
Flamingo-80B \cite{flamingo} & 80B & 14B & \underline{17.4} & \underline{35.6} & \textbf{40.7} \\
\hline 
PaLM2-VAdapter 1B (ViT-B) & 1.8B & 120M & 12.7 & 26.2 & 25.8 \\
PaLM2-VAdapter 1B  (ViT-L) & 2.0B & 120M & 14.0 & 18.6 & 28.3 \\
PaLM2-VAdapter 1B  (ViT-g) & 2.8B & 130M & 15.9 & 27.7 & 26.1 \\
PaLM2-VAdapter 8B  (ViT-g) & 10.8B & 130M & \textbf{19.6} & \textbf{40.5} & \underline{36.7}\\
\end{tabular}
 
\caption{\textbf{Zero-shot Video Question Answering.} The best result is \textbf{bolded} and second-best result is \underline{underlined}. Our model demonstrates state-of-the-art zero-shot multi-modal reasoning ability on videos.}
\vspace{-5mm}
    \label{tab:video_qa}
    
\end{table*}
\vspace{-5mm}

\subsection{Perceiver Resampler is Still Needed}
\label{sec:perceiver_is_needed}

In our first vision-language alignment stage (shown in \cref{fig:method_overview}b(i)), we follow CoCa~\cite{coca} to use an attentional pooler as the cross-attention module. This attentional pooler consists of a simple cross-attention layer and a LayerNorm layer for the final queried features.
Based on our observation of our in-depth empirical study with the perceiver resampler architecture (detailed in \cref{sec:strong_baseline_resampler}), we replace the attentional pooler with a 1-layer perceiver resampler to improve cross-modal alignment and achieve better performance, shown in \cref{tab:resampler_is_needed}. On the other hand, we observe that adding more layers of perceiver resampler does not lead to better performance with our adapter design which is contrary to the observation with vanilla perceiver resampler adaper. The empirical results show that a 1-layer perceiver resampler seems to be the best choice for cross-modality fusion in our proposed PaLM2-VAdapter.

\subsection{Visual Captioning}

\noindent\textbf{Image captioning.}  As detailed in \cref{tab:image_cap}, we evaluate the zero-shot image captioning performance on the COCO dataset~\cite{cococap}. Compared to the state-of-the-art AnyMAL model, our method shows comparable image captioning capability, but only requires  70\% parameters (10.8B vs. 15B), proving the effectiveness of our progressive alignment strategy. Additionally, the scalability of our PaLM2-VAdapter is evidenced through the vision encoder scaling experiment (from ViT-B to ViT-g),  indicating that a more powerful vision encoder correlates with enhanced image captioning performance. Qualitative examples are provided in \cref{fig:examples_caption} 
and \cref{sec:appendix_qualitative_examples}.

\noindent\textbf{Video captioning. } As detailed in \cref{tab:video_cap}, we evaluate the zero-shot video captioning performance on the MSRVTT and VATEX datasets~\cite{msrvtt,wang2019vatex}. Compared to the state-of-the-art Flamingo models, our method makes solid improvement on the VATEX benchmark but only requires  14\% parameters (10.8B vs. 80B). Similar to image captioning,  PaLM2-VAdapter still shows strong scalability when the vision encoder is scaled up. Moreover, scaling up language model also improves  video captioning performance, indicating that a larger language model lead to stronger ability to understand sequential visual information of videos. Qualitative examples are provided in \cref{fig:examples_caption} and \cref{sec:appendix_qualitative_examples}.

\vspace{-4mm}
\subsection{Visual Question Answering}

\begin{figure*}[!t]
    \centering
    \includegraphics[width=\linewidth]{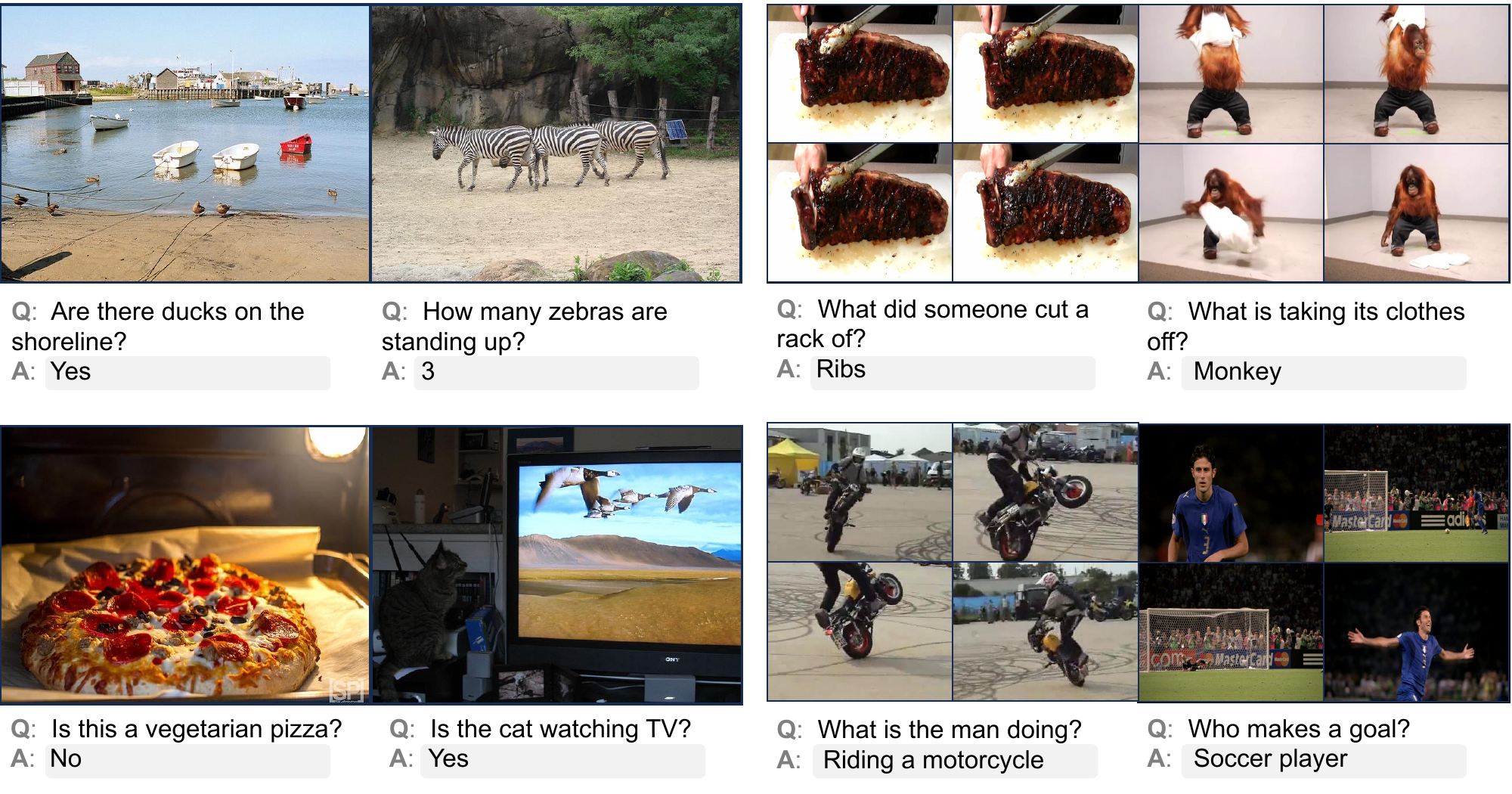} 
     
    \caption{\textbf{Qualitative examples of Visual Question Answering.} \textbf{Left:} Image question answering on the VQAv2 dataset. \textbf{Right:} video question answering on the MSVD-QA dataset.}
    \label{fig:examples_qa}

\end{figure*}

\noindent\textbf{Image question answering.} As detailed in \cref{tab:image_qa}, we evaluate the zero-shot image question answering performance on the VQAv2, TextVQA, VizWiz, and OKVQA datasets~\cite{vqav2,textvqa,bigham2010vizwiz,okvqa}. Compared to the state-of-the-art IDEFICS models, our method shows comparable image question answering ability but only requires 14\% parameters (10.8B vs. 80B), proving the effectiveness of our progressive alignment strategy. PaLM2-VAdapter shows very strong scalability - always achieving better performance when the vision encoder and LLM decoder are scaled up. Qualitative examples are provided in \cref{fig:examples_qa} and \cref{sec:appendix_qualitative_examples}.

\noindent\textbf{Video question answering.} As detailed in \cref{tab:video_qa}, we evaluate the zero-shot video question answering performance on the MSRVTT-QA, MSVD-QA and iVQA datasets~\cite{msrvtt, msvd_qa, yang2021justask}. Compared to the state-of-the-art Flamingo models, our method shows state-of-the-art video question answering ability but only requires 14\% parameters (10.8B vs. 80B), proving the remarkable effectiveness of our method. The results also justify the strong scalability of PaLM2-VAdapter.
Qualitative examples are provided in \cref{fig:examples_qa} and \cref{sec:appendix_qualitative_examples}.

\section{Limitation \& Discussion}
\label{sec:limitation_discussion}

Our PaLM2-VAdapter makes a significant improvement in efficiency, operating with substantially fewer parameters and much less training cost. However, its alignment process encounters challenges as the LLM decoder scales,  just like other large vision-language models. The key of this challenge lies in ensuring visual embeddings seamlessly transition into the scaled-up LLMs' input representation space. A potential solution involves the direct quantization of visual embeddings into language tokens, leveraging the shared LLM codebook across models of varying sizes for zero-shot transferability. So, here comes the question: \textbf{Can the visual embeddings be ``translated" to words?}

\begin{minipage}{0.4\textwidth}
To answer this question, we conduct a study to see if the visual embeddings output by the adapter can easily be ``translated" to a sequence of words and then used as the prefix for the LLM decoder. We introduce a fully-connected layer (FC layer) after the adapter and use the gumel-softmax operation~\cite{gumbel} to quantize the visual embeddings. 
\end{minipage}
\hfill
\begin{minipage}{0.55\textwidth}
\centering
\tablestyle{1pt}{1.1}
\begin{tabular}{ccc|c}
\multirow{2}{*}{Setting} & \multirow{2}{*}{Softmax Temp.} & \multirow{2}{*}{Temp. Decay} & \textbf{COCO} \\
                         &                              &                              & CIDEr         \\\shline
Baseline             & -                    & -                    & 44.1          \\ \hline
Gumbel-Softmax       & 1.0                    & -                    & 0             \\
Gumbel-Softmax       & 2.0                    & -                    & 13.1          \\
Gumbel-Softmax       & 2.0                    & Exponential$^{*}$                 & \textbf{15.3}
\end{tabular}

\captionof{table}{\textbf{Quantize the visual embeddings to words.} The baseline is aligned with image-text pairs. $^{*}$: the gumbel-softmax temperature is exponential decayed. }
\label{tab:quantize_to_words}
\end{minipage}

The output logits from the FC layer correspond to the words of the LLM codebook and the word with highest logit will be assigned to the corresponding visual token. As shown in \cref{tab:quantize_to_words}, the gumbel-softmax operation is very hard to train. We explore a lot of hyper-parameters to make the training  stable, however, the best result we got is just 15.3 CIDEr score on the COCO captioning dataset (shown in the last line), with the softmax temperature set to 2.0 and exponentially decayed. Compared to the baseline whose visual embeddings is not quantized, there is a huge performance drop when the visual embeddings are quantized to the words of LLM codebook. 

This implies that the visual embeddings might share the same representation space of LLM codebook but cannot be ``translated" to words with simple matching. We believe this is an interesting direction for future exploration: make the encoder and adapter have zero-shot scalability to larger LLMs.

\section{Conclusion}

In this paper, we propose PaLM2-VAdapter, which uses a tiny language model with progressive training strategy to effectively align vision encoders and LLMs.  Demonstrating exceptional zero-shot generalization capabilities across diverse vision-language tasks, PaLM2-VAdapter marks a significant stride in efficiency, operating with substantially fewer parameters than existing models.

Our contributions extend beyond mere technical enhancements in Large Vision-Language Models (LVLMs). We establish a simple but effective framework for future research in vision-language alignment, fostering advancements in multi-modal integration.  Morevover, the PaLM2-VAdapter's success in combining vision and language modality paves the way for further explorations, potentially revolutionizing various applications incorporating more modalities (\eg, audio, pose, ...). Our findings highlight the critical role and vast potential of adapter training strategy in the rapidly evolving domain of multi-modal alignment.

\bibliography{reference}

\clearpage
\appendix

\section{Implementation Details.}
\label{sec:imple_details}
\cref{tab:training_details} provides the detailed training recipe regarding to the hyper-parameters. We use 512 TPU v3 for training all the experiments. The stage-1 alignment takes 8 hours and stage-2 alignment takes around 20 hours. 
\begin{table}[ht]
    
	\centering
	\large
	\tablestyle{3pt}{1.1}
 \begin{tabular}	{c | c  c }
        Hyperparameter & \multicolumn{2}{c}{Setting} \\
        \shline
        Warmup steps & \multicolumn{2}{c}{1000}\\
        Learning rate & \multicolumn{2}{c}{5e-4} \\
        Batch size & \multicolumn{2}{c}{2048} \\
        AdamW $\beta$ & \multicolumn{2}{c}{(0.9,0.999)} \\
        Weight decay & \multicolumn{2}{c}{0.0001}\\
        Image resolution & \multicolumn{2}{c}{288}\\
        Patch size & \multicolumn{2}{c}{18}\\
        Prompt & \multicolumn{2}{c}{``Describe the following: $<$visual tokens$>$ :''}\\
        LLM decode mode & \multicolumn{2}{c}{Greedy}\\   
	\end{tabular}
	\caption
	{
        \textbf{Training settings for vision-language alignment.}
	}
	\label{tab:training_details}
\end{table}

\begin{table}[ht]
\centering
\large
\tablestyle{8pt}{1.3}
\begin{tabular}{c|c|l}
Benchmark   & Task Type                                 & \multicolumn{1}{c}{Prompt Template}                                                                                                                                               \\ \shline
COCO      & \multirow{3}{*}{Image \& Video Captioning }                         & \multirow{3}{*}{Describe the following $<$visual tokens$>$ :}                                                                                                                         \\
MSRVTT    &          &                                                                                                                                                                                   \\
VATEX     &                                           &                                                                                                                                                                                   \\ \hline
VQAv2     & \multirow{4}{*}{Image Question Answering} & \multirow{4}{*}{\shortstack[l]{Answer the question given the images. \\ Given $<$visual tokens$>$. \\ Question : $<$question$>$ ?\\ Answer:}}                     \\
TextVQA   &                                           &                                                                                                                                                                                   \\
Vizwiz    &                                           &                                                                                                                                                                                   \\
OKVQA     &                                           &                                                                                                                                                                                   \\ \hline
MSRVTT-QA & \multirow{3}{*}{Video Question Answering} & \multirow{3}{*}{\shortstack[l]{Answer the question given the images. \\ Given $<$visual tokens$>$. \\ Question : $<$question$>$ ?\\ Answer in exactly one word:}} \\
MSVD-QA   &                                           &                                                                                                                                                                                   \\
iVQA      &                                           &                                                                                                                                            
\end{tabular}
\caption{\textbf{ Prompt templates used for the Visual Captioning and Question Answering benchmarks.}}
\vspace{-1mm}
\end{table}

\section{Zero-Shot Generalization to VQA tasks}
\label{sec:zero_shot_vqa}

Following Flamingo~\cite{flamingo}, we use several pseudo samples from the downstream VQA tasks as the prompt prefix context (4 examples are used in our experiments). All the pseudo examples in the prompt are randomly selected from the training set and just include questions and answers (without image). An example zero-shot VQA prompt with two pseudo samples is:

\begin{center}
    \begin{lstlisting}
            Answer the question given the images.
            
            Given
            Question: Is the river clear?
            Answer: yes
            
            Given
            Question: Is this arctic or desert?
            Answer: desert
            
            Given  <visual embeddings>
            Image question: Where is he cooking? Answer: 
\end{lstlisting}
\end{center}

\section{Additional Qualitative Examples}
\label{sec:appendix_qualitative_examples}
\begin{figure}[h]
    \centering
    \includegraphics[width=0.9\linewidth]{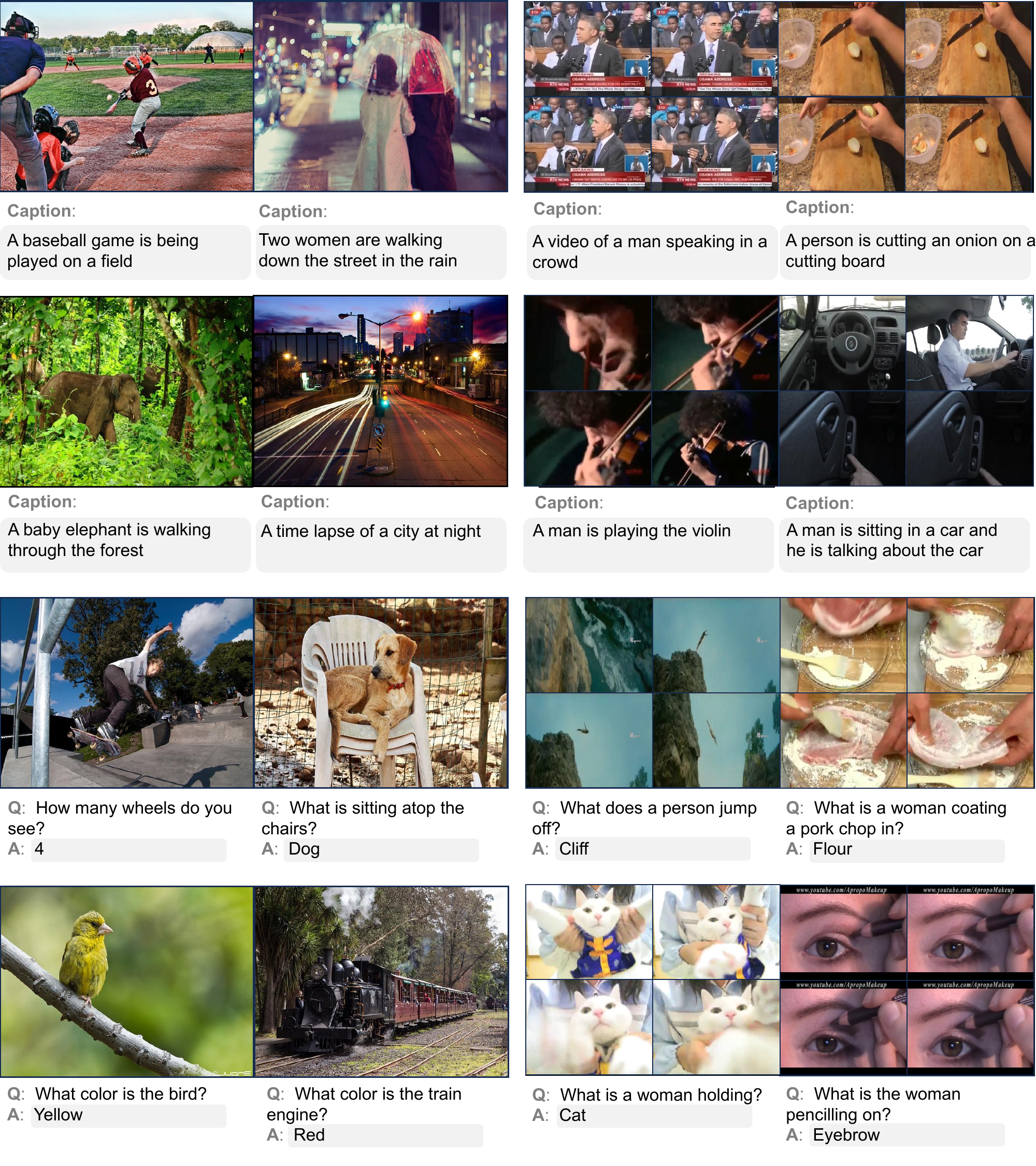}
    \caption{\textbf{Additional qualitative examples.} \textbf{Top left:} Image captioning on the COCO dataset. \textbf{Top right:} Video captioning on the MSRVTT dataset. \textbf{Bottom left:} Image question answering on the VQAv2 dataset. \textbf{Bottom right:} Video question answering on the MSVD-QA dataset.}
    \label{fig:qualitative_examples_appendix}
\end{figure}

\cref{fig:qualitative_examples_appendix} shows additional qualitative examples regarding to image and video tasks (\ie, captioning and QA). Our models demonstrate strong vision-language understanding ability and reasoning ability.

\section{Compatibility to CLIP Vision Encoder}
\label{sec:ablation_clip_encoder}

We conduct an additional experiment with a vision encoder pretrained with contrastive loss (same as CLIP) on the same image-text paired data used in CoCa. The table below clearly shows that our method leads to consistent improvements, whether we use ViT pre-trained with CLIP or ViT pre-trained with CoCa. This is in comparison to the state-of-the-art adapter design, known as the perceiver resampler.
\begin{table}[h]
\centering
\tablestyle{2pt}{1.3}
\begin{tabular}{ccccccc}
Vision Enc. & ViT Pretrain & LLM & Adapter & \textbf{COCO} Cap. & \textbf{MSRVTT} Cap. & \textbf{VQAv2} \\ \shline
ViT-B & CLIP~\cite{clip} & PaLM2-1B & Perceiver Resampler & 75.6 & 37.6 & 43.2 \\ 
ViT-B & CLIP~\cite{clip} & PaLM2-1B & PaLM2-VAdapter & 81.7 & 41.1 & 48.9 \\ 
ViT-B & CoCa~\cite{coca} & PaLM2-1B & Perceiver Resampler & 81.4 & 38.2 & 52.5 \\ 
ViT-B & CoCa~\cite{coca} & PaLM2-1B & PaLM2-VAdapter & \textbf{83.0} & \textbf{42.1} & \textbf{53.8}\\ 
\end{tabular}
\caption{\textbf{Generalization to other encoders.} Our PaLM2-VAdapter can also generalize to other vision encoders like CLIP~\cite{clip}.}
\label{tab:clip_encoder}
\end{table}

\section{Comparison to MLP adapter}

We also experiment with another baseline using the MLP architecture in the style of LLaVA v1.5~\cite{llava} (all the same experiment settings except using a different adapter), detailed in \cref{tab:mlp_adapter}. As shown in the following table, the MLP baseline performs much lower performance than our PaLM2-VAdapter.

\begin{table}[!h]
\centering
\tablestyle{2pt}{1.3}
\begin{tabular}{cccccc}
Vision Enc. & LLM & Adapter & \textbf{COCO} Cap. & \textbf{MSRVTT} Cap. & \textbf{VQAv2} \\ \shline
CoCa-B & PaLM2-1B & MLP & 80.3 & 41.2 & 43.4 \\ 
CoCa-B & PaLM2-1B & Perceiver Resampler & 81.4 & 38.2 & 52.5 \\ 
CoCa-B & PaLM2-1B & PaLM2-VAdapter & \textbf{83.0} & \textbf{42.1} & \textbf{53.8} \\ 
\end{tabular}
\caption{\textbf{Comparision to MLP Adapter.} Our PaLM2-VAdapter beats the commonly used MLP adapter (\eg, LLaVA~\cite{llava}) by a large margin.}
\label{tab:mlp_adapter}
\end{table}

\section{Broader Impact}
\label{sec:broader_impact}

This work presents a method to build vision language adapters effectively and efficiently. It fits in the broader context of large vision language models and share many of the benefits and issues of such models. The advancements in vision language models enable many useful applications across various fields.  However, it is crucial to acknowledge potential biases and ethical implications in the models, especially because the models utilizes pre-trained checkpoints and datasets and thus inherits such issues.  Research directions including mitigating biases in training data, improving algorithmic fairness and privacy-preserving techniques are becoming extremely vital to explore in order to address these harmful issues and benefit the broader community.

\end{document}